\documentclass{article} 
\usepackage{iclr2022_conference,times}

\usepackage{amsmath,amsfonts,bm}

\def\eqref#1{equation~\ref{#1}}

\def\1{\bm{1}}

\DeclareMathAlphabet{\mathsfit}{\encodingdefault}{\sfdefault}{m}{sl}
\SetMathAlphabet{\mathsfit}{bold}{\encodingdefault}{\sfdefault}{bx}{n}

\usepackage{hyperref}
\usepackage{array}
\usepackage{url}
\usepackage{graphics}

\begin{document}

\title{The \textit{African Stopwords} project: curating stopwords for African languages}


\author{Chris Emezue, Hellina Nigatu, Cynthia Thinwa, Helper Zhou, Shamsuddeen Muhammad, \And Lerato Louis,  Idris Abdulmumin, Samuel Oyerinde, Benjamin Ajibade,  Olanrewaju Samuel,  \And Oviawe Joshua, Emeka Onwuegbuzia,  Handel Emezue, Ifeoluwatayo A. Ige, \And Atnafu Lambebo Tonja , Chiamaka Chukwuneke, Bonaventure F.P. Dossou, Naome A. Etori\And Mbonu Chinedu Emmanuel, Oreen Yousuf, Kaosarat Aina, Davis David\\ 
Masakhane\\
}

%

\newcommand{\fix}{\marginpar{FIX}}
\newcommand{\new}{\marginpar{NEW}}

 \iclrfinalcopy 

\maketitle

\begin{abstract}
Stopwords are fundamental in Natural Language Processing (NLP) techniques for information retrieval.  One of the common tasks in preprocessing of text data is the removal of stopwords. Currently, while high-resource languages like English benefit from the availability of several stopwords, low-resource languages, such as those found in the African continent, have none that are standardized and available for use in NLP packages. Stopwords in the context of African languages are understudied and can reveal information about the crossover between languages. The \textit{African Stopwords} project aims to study and curate stopwords for African languages. In this paper, we present our current progress on ten African languages as well as future plans for the project.

\end{abstract}

\section{Introduction and Motivation}



When analysing text data and building various NLP models, stopwords might not add much value to the meaning of the document~\citep{Singh2019} depending on the NLP task (like text classification). Words such as articles and some verbs are usually considered stopwords because they don’t usually determine the context or the true meaning of a sentence -- they are words that can be removed without any negative consequences to the final NLP model training. 
Key to note also is that the removal of stopwords could improve a model training time owing to the reduced data size: the model will improve efficiency due to the reduced number of tokens involved in the training process~\citep{Singh2019}. 


However, stopword removal is highly dependent on the language, domain and task~\citep{Vallantin2019}. Therefore, the use of one ‘standard’ stopword list is problematic because it ignores the domain-knowledge specificity of stopwords~\citep{Lo2005AutomaticallyBA} and because it is language-specific~\citep{chineseStopwords}. Researchers~\citep{stopwords,doi:10.1177/016555159201800106,stopwordsNature} have shown that domain-specific and language-specific stopwords can make significant impact both in general tasks (like spam filtering, caption generation, language classification and auto-tag generation) and in domain-specific tasks (like NLP in the medical field or with the Chinese language\citep{chineseStopwords}). This shows the need for packages or techniques that can be utilised to effectively remove stopwords and enable building of effective NLP models. 

There are many available libraries for stopwords removal including Natural Language Toolkit(NLTK)~\citep{nltk}, spaCy~\citep{spaCy}, Gensim~\citep{gensim}, among others. Notwithstanding the ubiquity of stopwords packages in popular languages like English, Spanish, German, they do not support any African language (to the best of our knowledge). In this light, the main aim of our project is to curate stopwords for various African languages in order to positively contribute towards the advancement of natural language processing for African languages.


\section{\textit{African Stopwords}}
The \textit{African Stopwords} project aims to systematically gather stopwords for African languages -- starting with 13 African languages. 

\subsection{Current Progress}
Through the help of contributors, we have gathered some stopwords for the 13 focus African languages in Table~\ref{focusLanguages}.
We use a Github repository to host our code, contribution guides, curated stopwords, and linguistic discussions where there is controversy over potential stopwords candidates for a language.


\begin{table}[h!]
\centering
\renewcommand{\arraystretch}{1.3}
\resizebox{\columnwidth}{!}{%
\begin{tabular}{|m{2.3cm}|l|p{4.4cm}|p{0.2cm}|m{2cm}|l|p{4.4cm}|} 
  \hline
  \textbf{Lang.} & \textbf{\#} & \textbf{Source} & & \textbf{Lang.} & \textbf{\#} & \textbf{Source} \\ \cline{1-3}\cline{5-7}
  Afrikaans & 51 & \citep{tatman} & & Yoruba & 60 & \citep{tatman} \\ \cline{1-3}\cline{5-7}
  Hausa & 322 & \citep{tatman,hausa_stopwords} & & isiZulu & 29 & \citep{tatman} \\ \cline{1-3}\cline{5-7}
  Nigerian \mbox{Pidgin} & 34 & \citep{naijasenti} & & kiSwahili & 103 & \citep{tatman,davis} \\ \cline{1-3}\cline{5-7}
  Kirundi & 59 & \citep{niyongabo-etal-2020-kinnews} & & Igbo & - & - \\ \cline{1-3}\cline{5-7}
  Kinyarwanda & 80 & \citep{niyongabo-etal-2020-kinnews} & & Shona & - & - \\ \cline{1-3}\cline{5-7}
  Somali & 30 & \citep{tatman} & & Amharic & - & - \\ \cline{1-3}\cline{5-7}
  Sesotho & 31 & \citep{tatman} & & \multicolumn{3}{l|}{} \\ \hline
\end{tabular}
}
\caption{Focus African languages and current number of stopwords curated. `-' denotes languages designated for future work.}
\label{focusLanguages}
\end{table}

\subsection{future Work: Leveraging monolingual data}
One of the current directions of the project is in investigating the feasibility of harvesting online texts from multiple domains for curating better (language-specific and domain-relevant) African stopwords. There are numerous sources of monolingual data for African languages. Efforts for other languages have used monolingual knowledge sources such as
Brown Corpus \citep{10.1145/378881.378888,Maverick1969}, 20 newsgroup corpus~\citep{stopwordsNature}, books corpus~\citep{2010}, etc. In that respect, we plan to identify stopwords from monolingual data (of multiple domains) in the following steps:
\begin{enumerate}
    \item \textbf{Gather monolingual data:} The first part involves gathering a list of monolingual sources for the focus African languages. In order to ensure diverse, multi-domain stopwords, we will focus on getting data from many domains.
    \item \textbf{Using statistical methods to automatically identify candidate stopwords: }Research on stopwords identification have employed various statistical metrics, such as term-frequency-inverse-document-frequency (TF-IDF)~\citep{doi:10.1177/016555159201800106,Lo2005AutomaticallyBA}, entropy~\citep{2010}, information gain~\citep{10.1007/978-3-540-78646-7_22} and Kullback-Leibler divergence~\citep{Lo2005AutomaticallyBA}. We plan to use these statistical methods too to automatically identify candidate stopwords.
    \item \textbf{Human evaluation: }As a final step in ensuring the automatically curated stopwords are actually in line with the language, we will employ a number of human evaluators to review our stopwords. Only the stopwords that pass the evaluation will be published.
\item \textbf{Open-sourcing the stopwords:} Finally, we will either integrate the African stopwords into popular NLP processing toolkits (like NLTK) or create a separate Python package for it.

\end{enumerate}

\section{Conclusion}
Although the NLP community focuses on research in improvements of model development and data collection, there is still limited work on the inclusion of low resource languages in natural language processing toolkits. In this project, we set out to overcome one of those challenges by setting a framework to curate and accumulate stopwords for African languages. Finally we share our current progress on ten African languages as well as our plan for including more African languages. Our project is hosted at \url{https://github.com/masakhane-io/masakhanePreprocessor/tree/main/african-stopwords}.

\bibliography{iclr2022_conference}

\begin{thebibliography}{19}
\providecommand{\natexlab}[1]{#1}
\providecommand{\url}[1]{\texttt{#1}}
\expandafter\ifx\csname urlstyle\endcsname\relax
  \providecommand{\doi}[1]{doi: #1}\else
  \providecommand{\doi}{doi: \begingroup \urlstyle{rm}\Url}\fi

\bibitem[Abdulmumin \& Galadanci(2019)Abdulmumin and
  Galadanci]{hausa_stopwords}
Idris Abdulmumin and Bashir~Shehu Galadanci.
\newblock hauwe: Hausa words embedding for natural language processing.
\newblock In \emph{2019 2nd International Conference of the IEEE Nigeria
  Computer Chapter (NigeriaComputConf)}, pp.\  1--6, 2019.
\newblock \doi{10.1109/NigeriaComputConf45974.2019.8949674}.

\bibitem[Bird \& Loper(2004)Bird and Loper]{nltk}
Steven Bird and Edward Loper.
\newblock {NLTK}: The natural language toolkit.
\newblock In \emph{Proceedings of the {ACL} Interactive Poster and
  Demonstration Sessions}, pp.\  214--217, Barcelona, Spain, July 2004.
  Association for Computational Linguistics.
\newblock URL \url{https://aclanthology.org/P04-3031}.

\bibitem[David(2020)]{davis}
Davis David.
\newblock Swahili : News classification dataset, 2020.
\newblock URL \url{https://zenodo.org/record/5514203}.

\bibitem[Fox(1989)]{10.1145/378881.378888}
Christopher Fox.
\newblock A stop list for general text.
\newblock \emph{SIGIR Forum}, 24\penalty0 (1–2):\penalty0 19–21, sep 1989.
\newblock ISSN 0163-5840.
\newblock \doi{10.1145/378881.378888}.
\newblock URL \url{https://doi.org/10.1145/378881.378888}.

\bibitem[Gerlach et~al.(2019)Gerlach, Shi, and Amaral]{stopwordsNature}
Martin Gerlach, Hanyu Shi, and Lu{\'i}s A.~Nunes Amaral.
\newblock A universal information theoretic approach to the identification of
  stopwords.
\newblock \emph{Nature Machine Intelligence}, 1\penalty0 (12):\penalty0
  606--612, Dec 2019.
\newblock ISSN 2522-5839.
\newblock \doi{10.1038/s42256-019-0112-6}.
\newblock URL \url{https://doi.org/10.1038/s42256-019-0112-6}.

\bibitem[Honnibal et~al.(2020)Honnibal, Montani, Van~Landeghem, and
  Boyd]{spaCy}
Matthew Honnibal, Ines Montani, Sofie Van~Landeghem, and Adriane Boyd.
\newblock {spaCy: Industrial-strength Natural Language Processing in Python}.
\newblock 2020.
\newblock \doi{10.5281/zenodo.1212303}.

\bibitem[Lo et~al.(2005)Lo, He, and Ounis]{Lo2005AutomaticallyBA}
Rachel Tsz-Wai Lo, Ben He, and Iadh Ounis.
\newblock Automatically building a stopword list for an information retrieval
  system.
\newblock \emph{J. Digit. Inf. Manag.}, 3:\penalty0 3--8, 2005.

\bibitem[Makrehchi \& Kamel(2008)Makrehchi and
  Kamel]{10.1007/978-3-540-78646-7_22}
Masoud Makrehchi and Mohamed~S. Kamel.
\newblock Automatic extraction of domain-specific stopwords from labeled
  documents.
\newblock In Craig Macdonald, Iadh Ounis, Vassilis Plachouras, Ian Ruthven, and
  Ryen~W. White (eds.), \emph{Advances in Information Retrieval}, pp.\
  222--233, Berlin, Heidelberg, 2008. Springer Berlin Heidelberg.
\newblock ISBN 978-3-540-78646-7.

\bibitem[Maverick(1969)]{Maverick1969}
George~V. Maverick.
\newblock Computational analysis of present-day american english. henry
  ku{\v{c}}era , w. nelson francis.
\newblock \emph{International Journal of American Linguistics}, 35\penalty0
  (1):\penalty0 71--75, January 1969.
\newblock \doi{10.1086/465045}.
\newblock URL \url{https://doi.org/10.1086/465045}.

\bibitem[MONTEMURRO \& ZANETTE(2010)MONTEMURRO and ZANETTE]{2010}
MARCELO~A. MONTEMURRO and DAMIÁN~H. ZANETTE.
\newblock Towards the quantification of the semantic information encoded in
  written language.
\newblock \emph{Advances in Complex Systems}, 13\penalty0 (02):\penalty0
  135–153, Apr 2010.
\newblock ISSN 1793-6802.
\newblock \doi{10.1142/s0219525910002530}.
\newblock URL \url{http://dx.doi.org/10.1142/S0219525910002530}.

\bibitem[Muhammad et~al.(2022)Muhammad, Adelani, Ruder, Ahmad, Abdulmumin,
  Bello, Choudhury, Emezue, Salahudeen, Anuoluwapo, Jeorge, and
  Brazdil]{naijasenti}
Shamsuddeen~Hassan Muhammad, David~Ifeoluwa Adelani, Sebastian Ruder,
  Ibrahim~Said Ahmad, Idris Abdulmumin, Bello~Shehu Bello, Monojit Choudhury,
  Chris~Chinenye Emezue, Saheed~Abdullahi Salahudeen, Aremu Anuoluwapo,
  Al{\'{\i}}pio Jeorge, and Pavel Brazdil.
\newblock Naijasenti: {A} nigerian twitter sentiment corpus for multilingual
  sentiment analysis.
\newblock \emph{CoRR}, abs/2201.08277, 2022.
\newblock URL \url{https://arxiv.org/abs/2201.08277}.

\bibitem[Niyongabo et~al.(2020)Niyongabo, Hong, Kreutzer, and
  Huang]{niyongabo-etal-2020-kinnews}
Rubungo~Andre Niyongabo, Qu~Hong, Julia Kreutzer, and Li~Huang.
\newblock {KINNEWS} and {KIRNEWS}: Benchmarking cross-lingual text
  classification for {K}inyarwanda and {K}irundi.
\newblock In \emph{Proceedings of the 28th International Conference on
  Computational Linguistics}, pp.\  5507--5521, Barcelona, Spain (Online),
  December 2020. International Committee on Computational Linguistics.
\newblock \doi{10.18653/v1/2020.coling-main.480}.
\newblock URL \url{https://aclanthology.org/2020.coling-main.480}.

\bibitem[Sarica \& Luo(2020)Sarica and Luo]{stopwords}
Serhad Sarica and Jianxi Luo.
\newblock Stopwords in technical language processing.
\newblock \emph{CoRR}, abs/2006.02633, 2020.
\newblock URL \url{https://arxiv.org/abs/2006.02633}.

\bibitem[Singh(2019)]{Singh2019}
Shubham Singh.
\newblock How {To} {Remove} {Stopwords} {In} {Python} {\textbar} {Stemming} and
  {Lemmatization}.
\newblock \emph{Analytics Vidhya}, August 2019.
\newblock URL
  \url{https://www.analyticsvidhya.com/blog/2019/08/how-to-remove-stopwords-text-normalization-nltk-spacy-gensim-python/}.

\bibitem[Tatman(2017)]{tatman}
Rachael Tatman.
\newblock Stopword lists for african languages, Jul 2017.
\newblock URL
  \url{https://www.kaggle.com/rtatman/stopword-lists-for-african-languages}.

\bibitem[Vallantin(2020)]{Vallantin2019}
Lima Vallantin.
\newblock Why is removing stop words not always a good idea.
\newblock \emph{Medium}, June 2020.
\newblock URL
  \url{https://medium.com/@limavallantin/why-is-removing-stop-words-not-always-a-good-idea-c8d35bd77214}.

\bibitem[Wilbur \& Sirotkin(1992)Wilbur and
  Sirotkin]{doi:10.1177/016555159201800106}
W.~John Wilbur and Karl Sirotkin.
\newblock The automatic identification of stop words.
\newblock \emph{Journal of Information Science}, 18\penalty0 (1):\penalty0
  45--55, 1992.
\newblock \doi{10.1177/016555159201800106}.
\newblock URL \url{https://doi.org/10.1177/016555159201800106}.

\bibitem[Zou et~al.(2006)Zou, Wang, Deng, Han, and Wang]{chineseStopwords}
Feng Zou, Fu~Lee Wang, Xiaotie Deng, Song Han, and Lu~Sheng Wang.
\newblock Automatic construction of chinese stop word list.
\newblock In \emph{Proceedings of the 5th WSEAS International Conference on
  Applied Computer Science}, ACOS'06, pp.\  1009–1014, Stevens Point,
  Wisconsin, USA, 2006. World Scientific and Engineering Academy and Society
  (WSEAS).
\newblock ISBN 9608457432.

\bibitem[Řehůřek \& Sojka(2010)Řehůřek and Sojka]{gensim}
Radim Řehůřek and Petr Sojka.
\newblock {Software Framework for Topic Modelling with Large Corpora}.
\newblock Proceedings of LREC 2010 workshop New Challenges for NLP Frameworks,
  pp.\  45--50, Valetta, MT, 5 2010. University of Malta.
\newblock URL \url{http://is.muni.cz/publication/884893/en}.

\end{thebibliography}
\bibliographystyle{iclr2022_conference}


\end{document}